\documentclass{article} 
\usepackage{iclr2027_conference,times}

\usepackage{amsmath,amsfonts,bm}

\def\eqref#1{equation~\ref{#1}}

\def\1{\bm{1}}

\DeclareMathAlphabet{\mathsfit}{\encodingdefault}{\sfdefault}{m}{sl}
\SetMathAlphabet{\mathsfit}{bold}{\encodingdefault}{\sfdefault}{bx}{n}

\usepackage{hyperref}
\usepackage{url}
\usepackage{graphicx}
\usepackage{booktabs} 
\usepackage{multicol}
\usepackage{multirow}
\usepackage{listings}
\usepackage{algorithm}
\usepackage{algpseudocode}
\usepackage{amsmath,amssymb}
\usepackage{geometry}

\usepackage{listings}
\usepackage[most]{tcolorbox}
\usepackage{xcolor}
\usepackage{pifont}
\usepackage{wrapfig}
\usepackage{caption}
\usepackage{bbm}

\definecolor{promptbg}{RGB}{248,248,248}
\definecolor{promptframe}{RGB}{210,210,210}

\newtcblisting{promptbox}[1][]{
  enhanced,
  breakable,
  colback=promptbg,
  colframe=promptframe,
  boxrule=0.6pt,
  arc=2pt,
  left=7pt,
  right=7pt,
  top=6pt,
  bottom=6pt,
  fonttitle=\bfseries\small,
  title={#1},
  coltitle=black,
  colbacktitle=promptbg,
  attach boxed title to top left={
    xshift=6pt,
    yshift=-2mm
  },
  boxed title style={
    boxrule=0pt,
    colframe=promptbg,
    colback=promptbg,
    sharp corners
  },
  listing only,
  listing options={
    basicstyle=\ttfamily\small,
    breaklines=true,
    columns=fullflexible,
    keepspaces=true,
    showstringspaces=false,
    aboveskip=0pt,
    belowskip=0pt
  }
}

\usepackage{listings}
\usepackage[most]{tcolorbox}
\usepackage{xcolor}

\definecolor{boxbg}{RGB}{248,248,248}
\definecolor{boxframe}{RGB}{205,205,205}

\newtcblisting{examplebox}[1]{
  enhanced,
  breakable,
  listing only,
  colback=boxbg,
  colframe=boxframe,
  boxrule=0.6pt,
  arc=2pt,
  left=7pt,
  right=7pt,
  top=6pt,
  bottom=6pt,
  fonttitle=\bfseries\small,
  title={#1},
  coltitle=black,
  colbacktitle=boxbg,
  attach boxed title to top left={
    xshift=6pt,
    yshift=-2mm
  },
  boxed title style={
    boxrule=0pt,
    colframe=boxbg,
    colback=boxbg,
    sharp corners
  },
  listing options={
    basicstyle=\ttfamily\small,
    breaklines=true,
    columns=fullflexible,
    keepspaces=true,
    showstringspaces=false,
    aboveskip=0pt,
    belowskip=0pt
  }
}

\title{Learning More by Training Less: Uncertainty-Guided Progressive Contraction for Efficient Reasoning Post-Training}

\author{
    Suorong Yang\textsuperscript{1,2}, Jiaxin Wang\textsuperscript{2}, Jiaming Su\textsuperscript{3}, Chentao Yue\textsuperscript{4}, Furao Shen\textsuperscript{2 \dag}, Peng Ye\textsuperscript{5 \dag}, Soujanya Poria\textsuperscript{6 \dag}  \\
    \textsuperscript{1} National University of Singapore 
    \textsuperscript{2} Nanjing University 
    \textsuperscript{3} Fudan University \\
    \textsuperscript{4} The University of Sydney 
    \textsuperscript{5} The Chinese University of Hong Kong 
    \textsuperscript{6} Nanyang Technological University
}

\iclrfinalcopy 
\begin{document}

\maketitle

\begin{abstract}
Large language models have achieved remarkable advances in complex reasoning. However, scaling reasoning post-training with ever-larger corpora can substantially increase training cost without proportional performance gains.
In this study, we therefore ask: how far can reasoning supervision be compressed without sacrificing capability? 
To this end, we propose an uncertainty-guided progressive contraction framework that concentrates supervision and optimization on the reasoning trajectories most valuable for learning.
Specifically, rather than treating a teacher-generated chain-of-thought (CoT) as a single realized path, we construct \emph{uncertainty-profiled reasoning supervision} by building Monte Carlo reasoning trees around intermediate reasoning states to expose process-level uncertainty.
This uncertainty first contracts the original corpus into a compact teacher-informed pool. A second, learner-aware contraction then performs online routing during SFT, concentrating optimization according to the student's evolving learning state.
Together, the two stages progressively contract both supervision and optimization, substantially reducing the student training budget.
Experiments across diverse reasoning benchmarks and LLM families show that our method uses as little as 0.22\% of the optimization budget of full-data SFT, over 450$\times$ less training exposure, while maintaining or improving reasoning performance.
Notably, our framework generalizes well across heterogeneous student families, black-box teachers, and weak-to-strong transfer.
This indicates that it is not only a way to accelerate training but can also enhance reasoning capability.

\end{abstract}

\section{Introduction}
Large language models (LLMs) have made remarkable progress in complex reasoning, driven in part by increasingly capable pretrained models and post-training on ever-larger reasoning corpora~\citep{wei2023chainofthoughtpromptingelicitsreasoning,Guo_2025}.
A prevailing line of work is to scale supervision with more reasoning trajectories to train on~\citep{zelikman2022starbootstrappingreasoningreasoning,glm5team2026glm5vibecodingagentic}.
However, this scaling incurs substantial training cost without proportional performance gains, motivating a shift from simply increasing data volume toward identifying reasoning trajectories with greater learning value~\cite{he2026studentintheloopchainofthoughtdistillationgenerationtime,hou2026mattersonpolicydistillationperspective}.
Existing efforts broadly explore two complementary directions: constructing more effective reasoning supervision through stronger teacher generation and trajectory augmentation~\citep{Guo_2025,li-etal-2023-symbolic}, and improving the utilization of fixed corpora through filtering or routing based on correctness~\citep{slinko2026steprejectionfinetuningpractical,guha2025openthoughtsdatarecipesreasoning}, difficulty~\citep{tian2025deepdistillenhancingllmreasoning,huang2026tailoringcurriculumstudentcenteredreasoning}, or quality~\citep{li2026unifieddataselectionllm,just2026signalstepslocalscoring}.
While achieving promising results, neither direction is sufficient for extreme optimization efficiency.
As a result, achieving such efficiency requires coupling supervision construction with downstream allocation: first exposing where learning value resides, and then progressively concentrating optimization on it.

\begin{figure}[]
    \centering
    \includegraphics[width=0.9\linewidth]{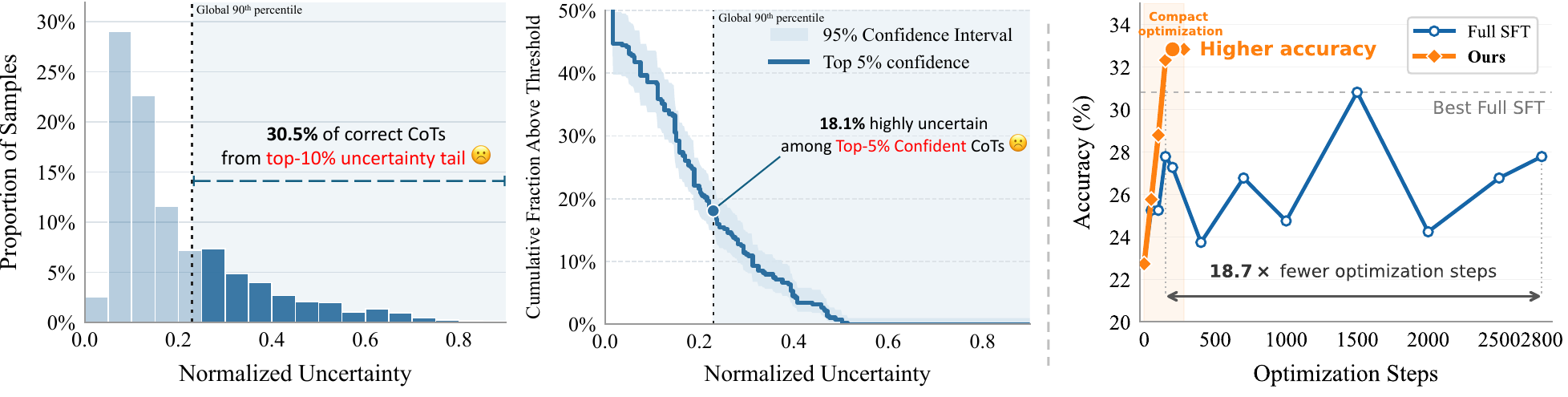}\vspace{-.2cm}
    \caption{\textbf{Correct and confident CoTs can still conceal fragile reasoning.} 
    \textit{Left}: Even among the top-5\% teacher-confident CoTs, \textbf{18.1\%} fall within the global top-10\% uncertainty tail.
    \textit{Mid}: Likewise, \textbf{30.5\%} of correct CoTs fall within the same uncertainty tail. 
    \textit{Right}: Motivated by this, we propose an uncertainty-guided progressive contraction framework. By progressively narrowing supervision from uncertainty-informative teacher reasoning to learner-relevant trajectories, we achieve competitive reasoning performance with 18.7$\times$ fewer post-training steps.}
    \label{fig:fig-1}
    \vspace{-.6cm}
\end{figure}

To expose learning-critical information, we exploit the internal structure of
reasoning.
In fact, reasoning is inherently non-uniform: a solution often hinges on a small number of intermediate states where plausible continuations diverge toward different outcomes, which we term \emph{reasoning forks}.
Such forks naturally manifest as process-level uncertainty, as alternative continuations from the same reasoning state may lead to both successful and unsuccessful solutions.
As shown in Fig.~\ref{fig:fig-1}, even highly confident or correct Chain of Thoughts (CoTs) can conceal substantial process-level uncertainty, suggesting that terminal confidence and correctness may mask fragile intermediate reasoning~\citep{
NEURIPS2022_639a9a17,zhao2026entropytrajectoryshapepredicts,
mao-etal-2026-confidence,ye2025limoreasoning}.
Consistent with prior work showing that uncertainty reveals local failures and reasoning boundary~\citep{kadavath2022languagemodelsmostlyknow,zhang2026stopfailoperationalcapability,mei2025reasoninguncertaintyreasoningmodels}, we therefore view process uncertainty as a behavioral signature of the reasoning capability boundary.

Motivated by this, we propose an uncertainty-guided progressive contraction framework for efficient reasoning post-training.
We first construct uncertainty-profiled reasoning supervision trajectories by building Monte Carlo Reasoning Trees (MCRTs) around intermediate states and estimating correct-answer reachability and branch-outcome uncertainty from sampled continuations.
Aggregated process uncertainty contracts the original corpus into a compact teacher-informed candidate pool along with uncertainty profiles.
During student SFT, however, the uncertainty-profiled CoTs remain learner-agnostic and do not reflect the student's evolving training state.
Thus, we introduce a second learner-aware contraction, which narrows the candidate set by routing optimization according to the student's evolving learning state.
Together, the two stages progressively concentrate supervision from
teacher-informative to learner-relevant trajectories, reducing both data usage
and, more importantly, the optimization required to learn from it.
As a result, our method uses as little as 0.22\% of the optimization budget of full-data SFT, \textbf{over 450$\times$ less} training exposure, while maintaining or improving reasoning performance.

Experimental results across diverse reasoning benchmarks and LLM architectures demonstrate that our method substantially accelerates SFT training while maintaining or improving reasoning performance.
Our framework presents strong cross-student transferability: reasoning
supervision distilled from the same Qwen3-8B~\citep{yang2025qwen3technicalreport} teacher generalizes effectively
across heterogeneous student families, including Qwen3.5-4B/9B~\citep{qwen3.5},
Gemma-4 E4B~\citep{gemmateam2026gemma4technicalreport}, Llama-3.2-3B-Instruct~\citep{grattafiori2024llama3herdmodels}, and
Mistral-7B-Instruct~\citep{jiang2023mistral7b}.
Importantly, the gains extend beyond efficiency: our 9B student achieves competitive reasoning performance with substantially larger reasoning models ranging from 30B to 671B parameters.
Notably, because uncertainty is characterized purely through observable
reasoning behavior, the same framework applies to \textbf{API-only black-box teachers} (e.g., DeepSeek-V4-Pro~\citep{deepseekai2026deepseekv4highlyefficientmilliontoken} and Kimi-K2.6~\citep{moonshot2026kimik26}) and teachers
substantially \textbf{smaller than their students}.
For instance, a Qwen3-1.7B teacher can improve a much stronger Qwen3.5-14B student, demonstrating effective reasoning transfer.


Our main contributions are as follows: 
\textbf{(1)} We introduce an uncertainty-guided progressive contraction framework for data-efficient reasoning post-training, jointly integrating supervision construction and downstream training allocation to progressively concentrate learning on high-value reasoning trajectories.
\textbf{(2)} We introduce uncertainty-profiled reasoning supervision, modeling MCRTs to expose process-level reasoning forks from observable model behavior without requiring model-internal access. 
\textbf{(3)} We develop learner-aware routing that further contracts the active training set during SFT according to the student's evolving learning state, enabling enhanced training acceleration.
\textbf{(4)} Across diverse benchmarks, model families, and teacher regimes, our method reduces reasoning optimization to as little as $<0.74\%$, while maintaining or improving reasoning performance, establishing a practical pathway toward data-efficient reasoning post-training for future research.

\section{The Proposed Method}
\subsection{Problem Formulation and Overview}
We consider reasoning distillation from a teacher model $T$ to a student model $S_\theta$ over a training set $\mathcal{D}={(x_i,a_i)}_{i=1}^{N}$.
For each problem $x_i$, the teacher generates a reasoning trajectory $\tau_i=(s{i,1},\ldots,s_{i,m_i},\hat a_i)$.
Rather than treating all verified CoTs as equally valuable supervision, we observe that their learning value depends on two factors: whether the
trajectory exposes informative structures in the teacher's reasoning space and
whether such information remains unresolved by the current learner.

Based on this observation, we introduce \emph{progressive supervision contraction}, which progressively narrows supervision through two stages: (1) teacher-side uncertainty contraction and (2) learner-aware optimization contraction.
\emph{In the first stage}, we treat each CoT as a backbone trajectory and construct an MCRT around its intermediate states to characterize process-level uncertainty.
Each trajectory is assigned a teacher-side uncertainty score $q_i^{T}$, and we
retain an uncertainty-rich candidate pool $\mathcal P_\rho$ with ratio $\rho$: $\mathcal P_\rho=\operatorname{Top}_{\rho}(\mathcal C;q_i^T)$.
Since this stage relies only on observable teacher behavior, $\mathcal P_\rho$
is student-agnostic and can be reused across different learners without
repeating uncertainty profiling.
\emph{In the second stage}, given the uncertainty-informed pool, we further allocate optimization according to the current learner state.
Specifically, we route training toward the $M$ trajectories with the largest
normalized response loss: $\mathcal H_M=
\arg\max_{\substack{
\mathcal H\subseteq\mathcal P_\rho,
|\mathcal H|=M}}
\sum_{i\in\mathcal H}\ell_i$,
where $\ell_i$ is obtained from standard forward passes without additional
optimization or auxiliary computation.
Subsequent training is performed exclusively on $\mathcal H_M$.
Together, the two stages reduce both the amount of data exposure and the optimization required for learning.
In practice, this reduces the student-side training budget to less than $1\%$
of vanilla full-set SFT while maintaining downstream reasoning performance.

\subsection{Uncertainty-Guided Reasoning and Profiling}\label{sec:teacher_uncertainty}
\paragraph{Uncertainty-Revealing Reasoning Elicitation.} 
For each problem $x_i$, we prompt the teacher to generate a structured reasoning trajectory $\tau_i=(s_{i,1},\ldots,s_{i,m_i},\hat a_i)$, where each $s_{i,t}$ corresponds to a semantically coherent reasoning step.
The trajectory induces prefix states $h_{i,t}=(x_i,s_{i,1:t}), t=1,\ldots,m_i,$ which serve as anchors for subsequent behavioral probing.
We treat $\tau_i$ as a backbone trajectory through the teacher's
reasoning space, and characterize the local reasoning stability around each $h_{i,t}$ through sampled continuations.

\paragraph{Monte Carlo Reasoning-Tree Profiling.}
Given a backbone trajectory $\tau_i$, we characterize the local stability of
each intermediate state $h_{i,t}$ by sampling $K$ independent prefix-conditioned continuations, $R_{i,t}^{(k)}\sim\pi_{\mathcal T}(\cdot\mid h_{i,t})$.
Let $\hat a_{i,t}^{(k)}$ denote the resulting final answer and
$Z_{i,t}^{(k)}=\mathbb{I}[\hat a_{i,t}^{(k)}=a_i]$ its correctness.
We model $Z_{i,t}^{(k)}\overset{\mathrm{i.i.d.}}{\sim}\operatorname{Bernoulli}(p_{i,t})$, where $p_{i,t}$ denotes the correct-answer reachability from $h_{i,t}$, and estimate it by $\hat p_{i,t}=K^{-1}\sum_{k=1}^{K}Z_{i,t}^{(k)}$.
Finite-sample concentration is provided in Appendix~\ref{app:mc_concentration}.

We then quantify local process uncertainty through the normalized Bernoulli variance 
\begin{equation}
    U_{i,t}
    =
    4p_{i,t}(1-p_{i,t})
    \in[0,1],
\end{equation}
with the plug-in estimator
\begin{equation}
    \widehat U_{i,t}
    =
    4\hat p_{i,t}(1-\hat p_{i,t}).
\end{equation}
 
This quantity directly measures branch-outcome instability.
For two independent continuations rooted at the same state, 
\begin{equation}
U_{i,t}
    =
    2\Pr\!\left(
        Z_{i,t}^{(1)}
        \neq
        Z_{i,t}^{(2)}
        \mid h_{i,t}
    \right).
\end{equation}
Hence, high $U_{i,t}$ identifies states whose plausible continuations diverge
in eventual correctness, rather than states that are simply difficult or
incorrect.

Notably, the entire profiling procedure is behavior-based: it operates only on
textual prefixes, sampled continuations, and verifiable outcomes, without
requiring logits, hidden states, gradients, or model parameters.
This makes the same uncertainty estimator directly applicable to heterogeneous
and API-only black-box teachers, provided that prefix-conditioned sampling and
outcome verification are available.

Finally, we aggregate state-wise uncertainty into the reasoning uncertainty profile $\mathcal Z_i = \left\{   (\hat p_{i,t},\widehat U_{i,t}) \right\}_{t=1}^{m_i}$ using the mean process uncertainty
 \begin{equation}
    \bar U_i
    =
    \frac{1}{m_i}
    \sum_{t=1}^{m_i}
    \widehat U_{i,t},
\end{equation}
which normalizes for trajectory length.
A larger $\bar U_i$ indicates that the realized trajectory more frequently passes through states whose alternative continuations disagree in eventual correctness.

\noindent \textbf{Uncertainty-Guided Supervision Contraction.}
We use the trajectory-level uncertainty $\bar U_i$ as the teacher-side
informativeness score, i.e., $q_i^{T}=\bar U_i$.
Given a candidate-pool ratio $\rho$, we perform the first supervision
contraction by retaining the verified trajectories with the largest
process uncertainty:
\begin{equation}
\mathcal P_\rho
=
\arg\max_{\substack{
    \mathcal S\subseteq\mathcal C,\\
    |\mathcal S|=\lfloor\rho|\mathcal C|\rfloor
}}
\sum_{i\in\mathcal S}\bar U_i .
\end{equation}
The resulting pool $\mathcal P_\rho$ concentrates supervision on verified trajectories that more frequently traverse states with unstable continuation outcomes, forming a compact teacher-informed candidate space for subsequent learner-aware optimization.

\subsection{Learner-Aware Data Routing}
The teacher-informed pool $\mathcal P_\rho$ reduces the amount of supervision
presented to the student before training.
We next perform the second contraction during SFT, routing optimization toward a small subset of trajectories according to the learner's evolving training state.
Concretely, we first perform a single probing epoch over $\mathcal P_\rho$.
For each trajectory $i\in\mathcal P_\rho$, let $\mathcal R_i$ denote the supervised response-token positions.
We derive the learner-side routing score $q_i^{S}$ using the
length-normalized response loss
\begin{equation}
    q_i^{S}
    =
    \ell_i
    =
    -\frac{1}{|\mathcal R_i|}
    \sum_{j\in\mathcal R_i}
    \log p_{\theta^{(w)}}
    \left(
        \tau_{i,j}\mid x_i,\tau_{i,<j}
    \right).
\end{equation}
A larger $\ell_i$ indicates that the corresponding teacher trajectory remains
less well captured by the current student and therefore receives higher routing
priority.

In this way, we contract the active training set by retaining the $M$ trajectories with the largest learner-side routing scores: \begin{equation}
    \mathcal H_M
    =
    \arg\max_{\substack{
        \mathcal H\subseteq\mathcal P_\rho,
        |\mathcal H|=M
    }}
    \sum_{i\in\mathcal H}\ell_i .
\end{equation}
Thus, after the probing epoch, subsequent SFT optimization is performed exclusively
over $\mathcal H_M$.
The routing scores are updated during training, allowing the active subset to adapt to the evolving learner state.
We provide a detailed algorithm pipeline in Appendix~\ref{app:algorithm-pipeline}.

\paragraph{Efficiency Analysis.}
Together, the two-stage contraction directly translates into a substantially smaller student optimization budget.
For instance, standard three-epoch SFT over $N$ trajectories processes $3N$ instances in total.
In contrast, our method trains on the largely contracted teacher-side pool only once
and restricts the subsequent two epochs to the learner-routed active set $3N
     \longrightarrow 
    |\mathcal P_\rho|+2|\mathcal H_M|
    =
    \lfloor\rho N_c\rfloor+2M,$ where $M \ll N$.
Accordingly, the relative optimization budget is 
\begin{equation}
    \mathcal B
    =
    \frac{\lfloor\rho N_c\rfloor+2M}{3N},
\end{equation}
where $N_c\leq N$ denotes the number of verified teacher trajectories.
This corresponds to training on only a small fraction of the trajectory exposures required by standard full-set SFT.
In practice, this reduces student-side training exposure to below \textbf{0.74\%} of vanilla full-set SFT, i.e., over \textbf{135$\times$ fewer} sample-level exposures, without sacrificing downstream reasoning performance.
We discuss more details in Appendix~\ref{app:training-cost-reduction}.

 \begin{table*}[]
\centering
\caption{Performance across competition mathematics, general mathematics, and science benchmarks using Qwen3.5-4/9B.
Training budget denotes cumulative student-side sample exposure relative to full-set SFT (100\%).
Overhead reports wall-clock GPU hours on a 2-H200-GPU server.
}
\label{tab:main_results}

\small
\setlength{\tabcolsep}{4.4pt}
\renewcommand{\arraystretch}{1.08}

\resizebox{\textwidth}{!}{
\begin{tabular}{@{}l cc|cccccc|cc|c@{}}
\toprule
& \multicolumn{2}{c|}{\textbf{Efficiency}}
& \multicolumn{6}{c|}{\textbf{Competition Math}}
& \multicolumn{2}{c|}{\textbf{General Math}}
& \multicolumn{1}{c}{\textbf{Science}}
\\
\cmidrule(lr){2-3}
\cmidrule(lr){4-9}
\cmidrule(lr){10-11}
\cmidrule(lr){12-12}

\textbf{Method}
& \textbf{Budget}
& \textbf{Overhead}
& \textbf{AIME24}
& \textbf{AIME25}
& \textbf{AIME26}
& \textbf{AMC}
& \textbf{HMMT25}
& \textbf{HMMT26}
& \textbf{MATH-500}
& \textbf{GSM8K}
& \textbf{GPQA-D} \\
\midrule

\multicolumn{12}{@{}l}{\textit{Qwen3.5-4B Student}} \\[2pt]

Full SFT
& 100\%
& 8.47
& 73.3
& 56.7
& 66.7
& 81.4
& 40.0
& 39.4
& 97.8
& 92.3
& 75.8 \\
\midrule

Ours ($\rho=30\%$)
& 1.91\%
& 0.07
& 86.7
& 83.3
& 86.7
& 96.3
& 76.7
& 69.7
& 98.4
& 96.4
& 77.8 \\

Ours ($\rho=10\%$)
& 0.74\%
& 0.03
& 86.7
& 83.3
& 83.3
& 94.8
& 76.7
& 78.8
& 98.0
& 96.2
& 77.3 \\

Ours ($\rho=5\%$)
& 0.45\%
& 0.02
& 86.7
& 73.3
& 80.0
& 93.3
& 70.0
& 69.7
& 97.8
& 96.3
& 77.3 \\

Ours ($\rho=1\%$)
& 0.22\%
& 0.01
& 90.0
& 83.3
& 86.7
& 94.0
& 76.7
& 69.7
& 98.4
& 96.3
& 77.8 \\

\midrule
\multicolumn{12}{@{}l}{\textit{Qwen3.5-9B Student}} \\[2pt]

Full SFT
& 100\%
& 9.23
& 90.0
& 86.7
& 90.0
& 93.3
& 73.3
& 63.6
& 97.8
& 96.7
& 78.8 \\
\midrule

Ours ($\rho=30\%$)
& 1.91\%
& 0.07
& 93.3
& 93.3
& 90.0
& 97.0
& 86.7
& 81.8
& 98.6
& 97.4
& 81.8 \\

Ours ($\rho=10\%$)
& 0.74\%
& 0.03
& 93.3
& 90.0
& 90.0
& 97.8
& 90.0
& 81.8
& 98.2
& 96.6
& 81.8 \\

Ours ($\rho=5\%$)
& 0.45\%
& 0.02
& 93.3
& 90.0
& 90.0
& 97.8
& 83.3
& 78.8
& 98.6
& 97.7
& 78.8 \\

Ours ($\rho=1\%$)
& 0.22\%
& 0.01
& 90.0
& 90.0
& 86.7
& 97.0
& 86.7
& 78.8
& 98.4
& 97.4
& 81.3 \\

\bottomrule
\end{tabular}
}
\end{table*}
\section{Experiment}\label{sec:experiment}
\subsection{Experimental Setup}
\textbf{Models and Training Data.}
We evaluate our framework across diverse student scales, architectures, and teacher regimes. Unless specified, we use Qwen3-8B~\cite{yang2025qwen3technicalreport} as the teacher, and Qwen3-14B and Qwen3.5-4B/9B~\cite{qwen3.5} as student models.
For cross-family generalization, we further evaluate Gemma-4-E4B-IT~\cite{gemmateam2026gemma4technicalreport}, Llama-3.2-3B-Instruct~\cite{grattafiori2024llama3herdmodels}, and Mistral-7B-Instruct~\cite{jiang2023mistral7b}.
Moreover, we consider API-only closed-source teachers, including DeepSeek-V4-Pro~\cite{deepseekai2026deepseekv4highlyefficientmilliontoken} and Kimi-K2.6~\cite{moonshot2026kimik26}, as well as a weaker Qwen3-1.7B teacher for
weak-to-strong transfer.
We train on DAPO-17K~\cite{yu2025dapoopensourcellmreinforcement}, while closed-source teacher experiments use the first 1000 samples from Nemotron-Math~\cite{du2025nemotronmathefficientlongcontextdistillation}.

\textbf{Training and Evaluation.} We train models with LoRA-based SFT~\cite{hu2021loralowrankadaptationlarge}. More training details, including optimization settings and configurations, are provided in Appendix~\ref{app:implementation_details}. 
We evaluate reasoning ability across competition mathematics, general mathematical reasoning, and scientific reasoning, including AIME24/25/26~\cite{aime25,aime26}, AMC~\cite{amc}, HMMT25~\cite{hmmt2025}, HMMT26~\cite{hmmt2026}, Math-500~\cite{math-500}, GSM8K~\cite{gsm8k}, and GPQA-Diamond (GPQA-D)~\cite{gpqa}.

\subsection{Main Results}
\textbf{Uncertainty-Guided Data-Efficient Training.}
As shown in Table~\ref{tab:main_results}, we evaluate reasoning distillation across competition mathematics, general mathematics, and scientific reasoning, using Qwen3-8B as a common teacher and Qwen3.5-4B/9B as students on DAPO-17k~\cite{}.
It can be seen that progressive contraction yields substantial reductions in both training budget and practical training costs while preserving or improving reasoning performance.
Across Qwen3.5-4B and 9B students, our method uses only \textbf{0.22--1.91\%} of the full-SFT optimization budget, reducing SFT time from 8.47/9.23 hours to only \textbf{0.01--0.07 hours}.
Under $\rho=1\%$, our method brings over \textbf{800$\times$} wall-clock acceleration, yet the resulting models remain competitive with or outperform full-set SFT across most benchmarks.
Importantly, this efficiency persists as the student scales from 4B to 9B,
highlighting its generality.
Moreover, performance remains remarkably stable under increasingly aggressive contraction, demonstrating that our progressive contraction effectively concentrates optimization on high-value supervision.

\begin{wrapfigure}[14]{r}{0.7\linewidth}
    \centering
    \vspace{-.2cm}
        \includegraphics[width=0.7\textwidth]{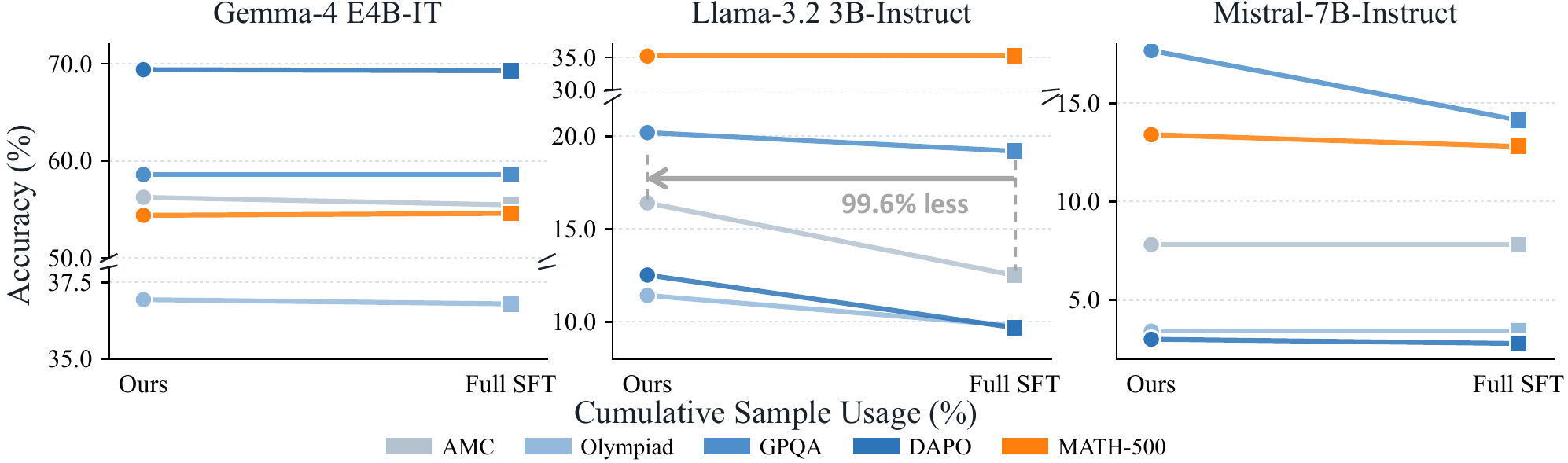}\vspace{-.1cm}
        \caption{Cross-family student generalization under a shared Qwen3-8B teacher. We evaluate Gemma, LLaMA, and Mistral students under the same teacher-side
    uncertainty profiling. The cumulative training budget is reduced by over 99.6\% compared to full-set SFT.}\label{fig:cross_arch_generalization}
\end{wrapfigure}
\textbf{Cross-Family Student Generalization.}
We next examine whether the optimization efficiency generalizes across distinct student model families.
Using the same Qwen3-8B teacher, we evaluate three heterogeneous students: Gemma-4-E4B-IT, LLaMA-3.2-3B-Instruct, and Mistral-7B-Instruct, covering different perspectives of reasoning capability.
As shown in Fig.~\ref{fig:cross_arch_generalization}, our method maintains or improves overall reasoning performance across all three student families while using only approximately \textbf{0.45\%} of the full-SFT student-side
optimization budget, corresponding to over \textbf{200$\times$ training acceleration}.
The consistent effectiveness under such aggressive contraction demonstrates that the efficiency gains are not specific to a particular student architecture. 
Importantly, once the teacher-side uncertainty profile is constructed from the same
Qwen3-8B teacher, it can be shared across students, highlighting strong cross-family transfer and effectiveness.

\begin{table}[]
\centering
\caption{Weak-to-strong reasoning transfer with a 1.7B teacher and Qwen3.5-4B and Qwen3-14B students with $\rho=5\%$ on DAPO-17K. Overhead is the wall-clock time (h) on a 2-H200-GPU server.}
\label{tab:weak-to-strong}
\setlength{\tabcolsep}{3.8pt}
\renewcommand{\arraystretch}{1.06}
\small

\begin{tabular}{@{}lcccccc@{}}
\toprule
\textbf{Model}
& \textbf{AIME24}
& \textbf{AIME25}
& \textbf{GPQA-D}
& \textbf{MATH-500}
& \textbf{GSM8K}
& \textbf{Overhead} \\
\midrule

Qwen3-1.7B (\textit{Weak teacher})
& 48.3 & 36.8 & 40.1 & 83.4 & 90.0 & - \\

\midrule

Qwen3.5-4B
& 73.8 & 65.6 & 76.2 & 98.0 & 96.1 & - \\[-1pt]

\quad \textbf{+ Ours (5\%)}
& \textbf{83.3}\textsuperscript{\scriptsize +9.5}
& \textbf{80.0}\textsuperscript{\scriptsize +14.4}
& \textbf{76.3}\textsuperscript{\scriptsize +0.1}
& \textbf{98.6}\textsuperscript{\scriptsize +0.6}
& \textbf{96.4}\textsuperscript{\scriptsize +0.2}
& +0.02h \\

\midrule

Qwen3-14B
& 76.0 & 67.3 & 61.5 & 96.0 & 89.8 & - \\[-1pt]

\quad \textbf{+ Ours (5\%)}
& \textbf{86.7}\textsuperscript{\scriptsize +10.7}
& \textbf{70.0}\textsuperscript{\scriptsize +2.7}
& \textbf{62.2}\textsuperscript{\scriptsize +0.7}
& 95.6\textsuperscript{\scriptsize $-0.4$}
& \textbf{95.5}\textsuperscript{\scriptsize +5.7}
& +0.10h \\

\bottomrule
\end{tabular}
\vspace{-.2cm}
\end{table}

\textbf{Weak-to-Strong Reasoning Transfer.}
Conventional distillation typically assumes a more capable teacher than the target student.
In contrast, we use Qwen3-1.7B as a weak teacher for the stronger Qwen3.5-4B and Qwen3-14B students.
As shown in Table~\ref{tab:weak-to-strong}, under $\rho=5\%$, the 1.7B teacher still provides effective uncertainty-guided supervision, yielding gains of up to 14.4\% while requiring merely $\leq0.1$ GPU hours of student training.
These results provide strong evidence for \emph{weak-to-strong reasoning
transfer}, suggesting that supervision value need not be determined solely by
teacher strength: uncertainty-guided profiling can expose locally informative
reasoning trajectories, while learner-aware routing concentrates training on
those that remain useful to the stronger student.

\textbf{Black-Box Distillation with Closed-Source Teachers.}
A practical advantage of our framework is that teacher-side uncertainty profiling is entirely behavior-based.
Thus, our method can be directly applied to API-only closed-source teacher models without access to their internal parameters or gradients.
Specifically, we use DeepSeek-V4-Pro and Kimi-K2.6 as closed-source teachers, paired with Qwen3.5-4B and Qwen3.5-9B, respectively.
We move beyond DAPO-17k and use the first 1000 samples from Nemotron-Math~\cite{du2025nemotronmathefficientlongcontextdistillation} to keep API querying practical.
As shown in Fig.~\ref{fig:closed-teacher-distillation}, uncertainty-guided
progressive contraction consistently improves over vanilla distillation across
both closed-source teachers and diverse reasoning benchmarks, with gains of up
to \textbf{13.4\%}.
Importantly, no modification to the uncertainty estimator is required when
moving from open-weight to API-only teachers: process uncertainty is recovered
solely from observable continuation behavior, highlighting its practicality.

\begin{table*}[]
\centering
\caption{
Comparison with larger reasoning models across various reasoning benchmarks.
\textbf{Bold} and \underline{underlined} values denote the best and
second-best results, respectively.
}
\label{tab:large_model_comparison}

\small
\setlength{\tabcolsep}{5.2pt}
\renewcommand{\arraystretch}{1.10}

\resizebox{.95\textwidth}{!}{
\begin{tabular}{@{}l c @{\hspace{8pt}} ccccccc@{}}
\toprule
\textbf{Model}
& \textbf{Params}
& \textbf{AIME24}
& \textbf{AIME25}
& \textbf{AIME26}
& \textbf{HMMT25}
& \textbf{HMMT26}
& \textbf{MATH-500}
& \textbf{GPQA-D} \\
\midrule

\multicolumn{9}{@{}l}{\textit{Larger Reasoning Models}} \\[2pt]

Gemma-4-it
& 12B
& --
& 72.9
& 77.5
& 63.3
& \underline{72.7}
& --
& 78.8 \\

Ministral-3-Reasoning-2512
& 14B
& 89.8
& 82.9
& 85.0
& 67.1
& 42.4
& --
& 71.2 \\

Phi-4-Reasoning-Plus
& 14B
& 81.3
& 68.4
& 73.6
& 50.3
& 51.5
& 83.0
& 68.9 \\

Nemotron-3-Nano
& 30B
& --
& 89.1
& \underline{89.9}
& \underline{84.6}
& --
& --
& 74.1 \\

GLM-4.5-Air
& 106B
& 89.4
& 83.3
& --
& 69.2
& --
& 98.1
& -- \\

Qwen3-235B-A22B-Thinking
& 235B
& 85.7
& \underline{92.3}
& --
& 83.9
& --
& 98.0
& 81.1 \\

LongCat-Flash-Thinking
& 560B
& \textbf{93.3}
& 90.6
& --
& 83.7
& --
& \textbf{99.2}
& \underline{81.5} \\

DeepSeek-R1-0528
& 671B
& \underline{91.4}
& 87.5
& --
& 79.4
& --
& 98.3
& 81.0 \\

GPT-5 Nano (high)
& N/A
& --
& 85.2
& 76.7
& 75.6
& 69.7
& --
& 71.2 \\

\midrule


\textbf{Ours-9B}
& \textbf{9B}
& \textbf{93.3}
& \textbf{93.3}
& \textbf{90.0}
& \textbf{90.0}
& \textbf{81.8}
& \underline{98.6}
& \textbf{81.8} \\
\bottomrule
\end{tabular}
}
\vspace{-.5cm}
\end{table*}

\textbf{Comparison with Substantially Larger Reasoning Models.}
As shown in Table~\ref{tab:large_model_comparison}, our compact 9B student remains highly competitive with reasoning models that are one to two orders of magnitude larger. 
Notably, this suggests that the supervision identified by progressive contraction not only improves optimization efficiency, but also translates into substantial capability gains for the resulting student.
Thus, our framework can produce a substantially stronger reasoner without relying on increased model scale.
\begin{figure*}
    \begin{minipage}{0.68\textwidth}
        \centering
        \includegraphics[width=0.99\textwidth]{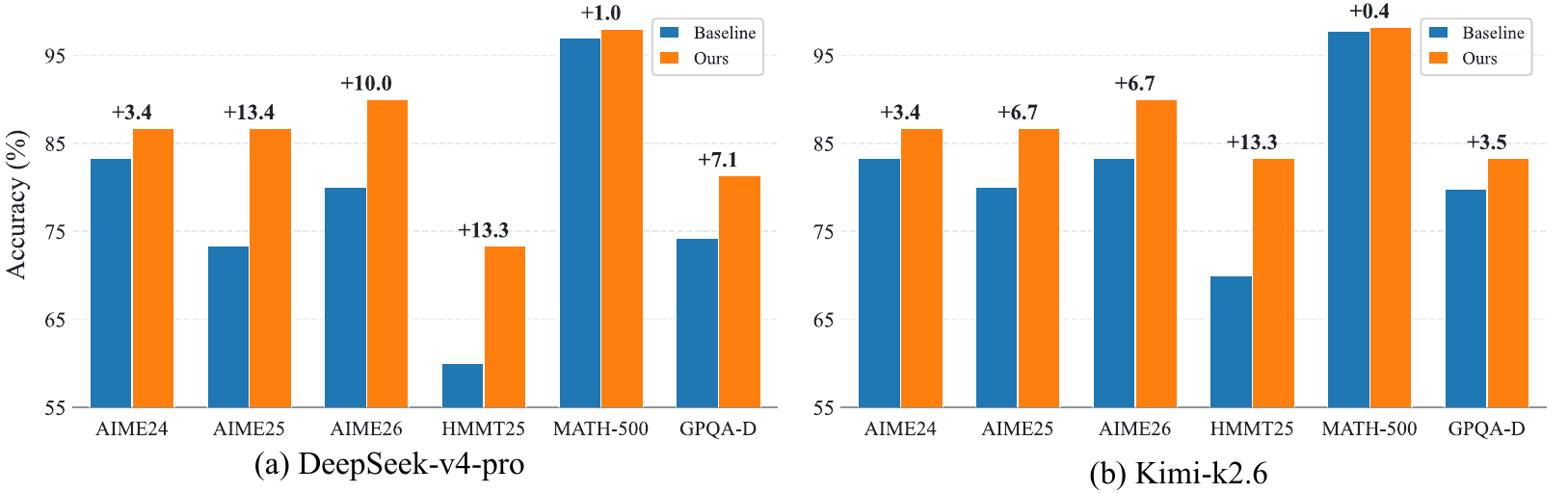}\vspace{-.2cm}
        \caption{Generalization to closed-source teachers using (a) DeepSeek-V4-Pro with Qwen3.5-4B, and (b) Kimi-K2.6 with Qwen3.5-9B.}
        \label{fig:closed-teacher-distillation}
    \end{minipage}
    \hfill
    \begin{minipage}{0.31\textwidth}
         \centering
    \captionof{table}{Uncertainty-based selection versus random selection under matched budgets using Qwen3.5-4B. We report average results across 7 benchmarks.} \vspace{.1cm}
    \label{tab:uncertainty_vs_random}
    \setlength{\tabcolsep}{4pt}
    \resizebox{\linewidth}{!}{
    \begin{tabular}{c|cccc}
        \toprule
        \textbf{Pool Ratio} 
        & \textbf{1\%} 
        & \textbf{5\%} 
        & \textbf{10\%} 
        & \textbf{30\%} \\
        \midrule
        Random 
        & 78.1 & 76.2 & 75.6 & 76.5 \\
        High-U 
        & \textbf{79.7} 
        & \textbf{80.6} 
        & \textbf{80.1} 
        & \textbf{78.7} \\
        $\Delta$ 
        & \textbf{+1.6} 
        & \textbf{+4.4} 
        & \textbf{+4.5} 
        & \textbf{+2.2} \\
        \bottomrule
    \end{tabular}
    }
    \end{minipage}
\end{figure*}

\subsection{Mechanism Analysis and Ablations}

\begin{table*}[t]
\centering
\caption{Matched-budget comparison with vanilla distillation under a 1\% training budget.}
\label{tab:matched_budget_1pct}
\setlength{\tabcolsep}{4.8pt}
\renewcommand{\arraystretch}{1.12}
\resizebox{.95\textwidth}{!}{
\begin{tabular}{@{}llccccccccc@{}}
\toprule
\textbf{Student}
& \textbf{Method}
& \textbf{AIME24}
& \textbf{AIME25}
& \textbf{AIME26}
& \textbf{AMC}
& \textbf{GPQA-D}
& \textbf{HMMT25}
& \textbf{HMMT26}
& \textbf{MATH500}
& \textbf{GSM8K} \\
\midrule

\multirow{2}{*}{Qwen3.5-4B}
& Vanilla Distill.
& 86.7
& 73.3
& 76.7
& 93.3
& \textbf{77.8}
& 73.3
& 66.7
& 97.0
& 96.2 \\

& \textbf{Ours}
& \textbf{90.0}
& \textbf{83.3}
& \textbf{86.7}
& \textbf{94.0}
& \textbf{77.8}
& \textbf{76.7}
& \textbf{69.7}
& \textbf{98.4}
& \textbf{96.4} \\
\midrule

\multirow{2}{*}{Qwen3.5-9B}
& Vanilla Distill.
& 86.7
& 86.7
& 83.3
& 95.5
& 80.3
& 80.0
& 63.6
& 97.8
& 97.0 \\

& \textbf{Ours}
& \textbf{90.0}
& \textbf{90.0}
& \textbf{86.7}
& \textbf{97.0}
& \textbf{81.3}
& \textbf{86.7}
& \textbf{78.8}
& \textbf{98.4}
& \textbf{97.4} \\

\bottomrule
\end{tabular}}
\vspace{-5pt}
\end{table*}

\begin{table*}[!htbp]
\centering
\caption{Effect of teacher-side uncertainty profiling and learner-aware routing on Qwen3.5-4B with $\rho=$10\%.}
\label{tab:teacher_learner_ablation}
\resizebox{.9\linewidth}{!}{
\begin{tabular}{cc|cccccc}
\toprule
\multicolumn{2}{c|}{Components}
& \multicolumn{6}{c}{Benchmark Accuracy (\%)} \\
\cmidrule(lr){1-2}
\cmidrule(lr){3-8}
Teacher Unc. & Learner Routing
& AIME24 & AIME25 & AIME26
& GPQA & HMMT25 & HMMT26 \\
\midrule

$\times$ & $\times$
& 83.3 & 73.3 & 76.7
& 75.8 & 63.3 & 54.5 \\

$\checkmark$ & $\times$
& 80.0 & 76.7 & 76.7
&  76.3 & 73.3 & 57.6 \\

$\times$ & $\checkmark$
& 83.3 & 76.7 & 80.0
& 76.3 & 73.3 & 63.6 \\

$\checkmark$ & $\checkmark$
& \textbf{86.7} & \textbf{83.3} & \textbf{83.3}
& \textbf{77.3} & \textbf{76.7} & \textbf{78.7} \\

\bottomrule
\end{tabular}
}
\vspace{-.3cm}
\end{table*}
\noindent \textbf{Teacher-Side Uncertainty Identifies High-Value Reasoning Trajectories.}
To evaluate whether uncertainty-guided reasoning provides more effective supervision than standard distillation, we prompt the same teacher to generate standard CoTs and train the students using the same optimization budget.
As shown in Table~\ref{tab:matched_budget_1pct}, our method consistently outperforms vanilla distillation on both Qwen3.5-4B and 9B, improving the average performance by 3.6\% and 4.0\%, respectively.
Because the student optimization budget is strictly matched, the
gains reflect more effective allocation of the optimization budget.

To isolate teacher-side uncertainty, we disable learner-aware routing and compare
uncertainty-based selection with matched-size random selection among the same
verified-correct trajectories.
As shown in Table~\ref{tab:uncertainty_vs_random}, uncertainty-based selection consistently improves average performance by 1.6\%-4.5\%.
This shows that final-answer correctness alone does not fully characterize
supervision value; process uncertainty further distinguishes more informative
reasoning trajectories among successful CoTs.

\noindent \textbf{Efficiency Comes from Better Optimization Budget Allocation.}
To better understand the efficiency of our method, as shown in Figure~\ref{fig:fig-1}, we track the optimization trajectories and compare them with the full SFT. 
It shows that our method surpasses the full SFT within only 150 steps, whereas full SFT requires about 1500 steps to reach its optimum, \textbf{corresponding to $10\times$ fewer optimization steps}.
These results show that progressive contraction substantially
increases the learning progress by concentrating optimization on high-value supervision, thereby improving
reasoning performance with dramatically less student training.



\textbf{Teacher Uncertainty and Learner Routing Are Complementary.}
We ablate both stages of progressive contraction.
As shown in Table~\ref{tab:teacher_learner_ablation}, each component improves accuracy, while their combination yields the best performance.
Importantly, both stages contribute along complementary axes: the former improves the informativeness of the candidate
supervision space, while learner-aware routing concentrates SFT updates on trajectories with greater student-specific learning demand.
Together, they provide a new perspective by concentrating increasingly effective supervision into an increasingly compact optimization set, enabling stronger reasoning performance and substantially higher learning efficiency simultaneously.
The substantially larger gain from their combination supports the central
design of progressive contraction: first preserve informative reasoning
supervision, and then allocate optimization according to training dynamics.
\section{Related Work}
\noindent \textbf{Uncertainty in LLM Reasoning and Learning.}
Uncertainty quantification has been studied as a means of characterizing the reliability and knowledge boundaries of LLMs~\citep{shorinwa2025survey,li-etal-2025-knowledge-boundary,ye2025mow}. 
Existing approaches estimate uncertainty through token probabilities, verbalized confidence, sampling-based disagreement, and semantic entropy~\citep{manakul-etal-2023-selfcheckgpt,farquhar2024detecting,luo2026your}.
Recent studies further show that stronger or longer reasoning does not
necessarily yield better uncertainty awareness, with modern reasoning models
often remaining overconfident or increasingly miscalibrated on incorrect
predictions~\citep{mei2025reasoninguncertaintyreasoningmodels,yeom2026epicar}.
Beyond final-answer confidence, process-level uncertainty has been modeled over
intermediate reasoning states through Monte Carlo continuation, confidence
probing, and calibration-aware supervision~\citep{wang2026process,park2026know,yan2026mur}.
For instance, RLCM~\citep{wang2026processsupervisionconfidencemargin} introduces confidence-aware process supervision and improves calibration while preserving
reasoning accuracy.
These works motivate treating uncertainty as a structured signal of
reasoning stability and capability boundaries throughout multi-step reasoning.~\citep{wang2026process,park2026know,yan2026mur}.

\noindent \textbf{Data-Efficient Distillation and Adaptive Data Routing.}
Knowledge distillation provides a data-centric mechanism for transferring capabilities from stronger teachers to smaller learners~\citep{hsieh2023distillingstepbystepoutperforminglarger,shridhar2023distillingreasoningcapabilitiessmaller,zhang2026questefficientreasoningdatacentric}.
For reasoning tasks, recent work increasingly distills intermediate rationales or long-form CoTs rather than final answers alone, enabling students to acquire richer reasoning behaviors~\citep{liu2026long,wu2026concise,wu2026enhancing}.
Beyond increasing the amount of reasoning supervision, growing evidence suggests that its effectiveness depends strongly on trajectory form and learner compatibility~\citep{yang2026reasoning,zhang2026skillawaredataselectionfinetuning}.
Accordingly, reasoning supervision has been adapted through trajectory filtering~\citep{yang2026reasoning}, prefix alignment~\citep{liu2026long}, difficulty-aware supervision~\citep{wu2026concise}, and student-conditioned trajectory selection~\citep{zhang2026skill,he2026studentintheloopchainofthoughtdistillationgenerationtime}.
In parallel, data selection and valuation methods allocate training computation toward more useful examples~\citep{yang2025reinforcement,opt-based-3}, using signals such as quality~\citep{wang2024greats,yang2025clip}, diversity~\citep{ccs}, gradient or influence information~\citep{yang2022dataset,wang2026opus}, and model-dependent learning dynamics~\citep{engstrom2024dsdm,wang2025nice}.
Representative approaches include DEITA~\citep{liu2024makes}, LESS~\citep{less}, Data Agent~\citep{yang2026data}, and RHO-Loss~\citep{mindermann2022prioritized}.
These methods can be broadly viewed as data routing: allocating limited optimization toward supervision with greater expected learning value.
\section{Conclusion}
We propose an uncertainty-guided progressive contraction framework that exposes
and concentrates learning value in reasoning supervision, substantially
accelerating reasoning post-training without sacrificing capability.
Across diverse benchmarks and architectures, our method consistently improves
data efficiency with strong generalization.
These results highlight that effective reasoning supervision depends not only
on scale, but also on how learning value is identified and allocated throughout training.
More broadly, we hope this work inspires further research on the shift from scaling reasoning data toward more efficient and accessible reasoning post-training, while providing a practical tool to reduce training burden and enable strong reasoning models under limited computational resources for the community.

\bibliography{iclr2027_conference}
\bibliographystyle{iclr2027_conference}

\end{document}